\documentclass[lettersize,journal]{IEEEtran}
\usepackage{amsmath,amsfonts}
\usepackage{algorithmic}
\usepackage{algorithm}
\usepackage{array}
\usepackage[caption=false,font=footnotesize,labelfont=rm,textfont=rm]{subfig}
\usepackage{textcomp}
\usepackage{stfloats}
\usepackage{url}
\usepackage{verbatim}
\usepackage{graphicx}
\usepackage{cite}
\usepackage{multirow}
\begin{document}

\title{GaussVid: Sparse-View Gaussian Splatting with 3D-Aware \\ Video Diffusion Priors}

\author{Xinhui Liu, Can Wang, Wei Jiang, Wei Wang, Dong Xu, \IEEEmembership{Fellow, IEEE} 
\thanks{X. Liu and C. Wang are with the School of Computing and Data Science, The University of Hong Kong, Hong Kong, China. E-mail: (xhliu01 and canwang)@hku.hk.}
\thanks{W. Jiang and W. Wang are with the Futurewei Technologies Inc, Santa Clara, CA, USA. E-mail: (wjiang, rickweiwang)@futurewei.com. }
\thanks{D. Xu is with the School of Computing and Data Science, The University of Hong Kong, Hong Kong, China. E-mail: dongxu@hku.hk. Corresponding Author.}
}
\markboth{Journal of \LaTeX\ Class Files,~Vol.~14, No.~8, August~2021}%
{Shell \MakeLowercase{\textit{et al.}}: A Sample Article Using IEEEtran.cls for IEEE Journals}


\maketitle

\begin{figure*}[!t]
  \centering
  \includegraphics[width=0.94\textwidth]{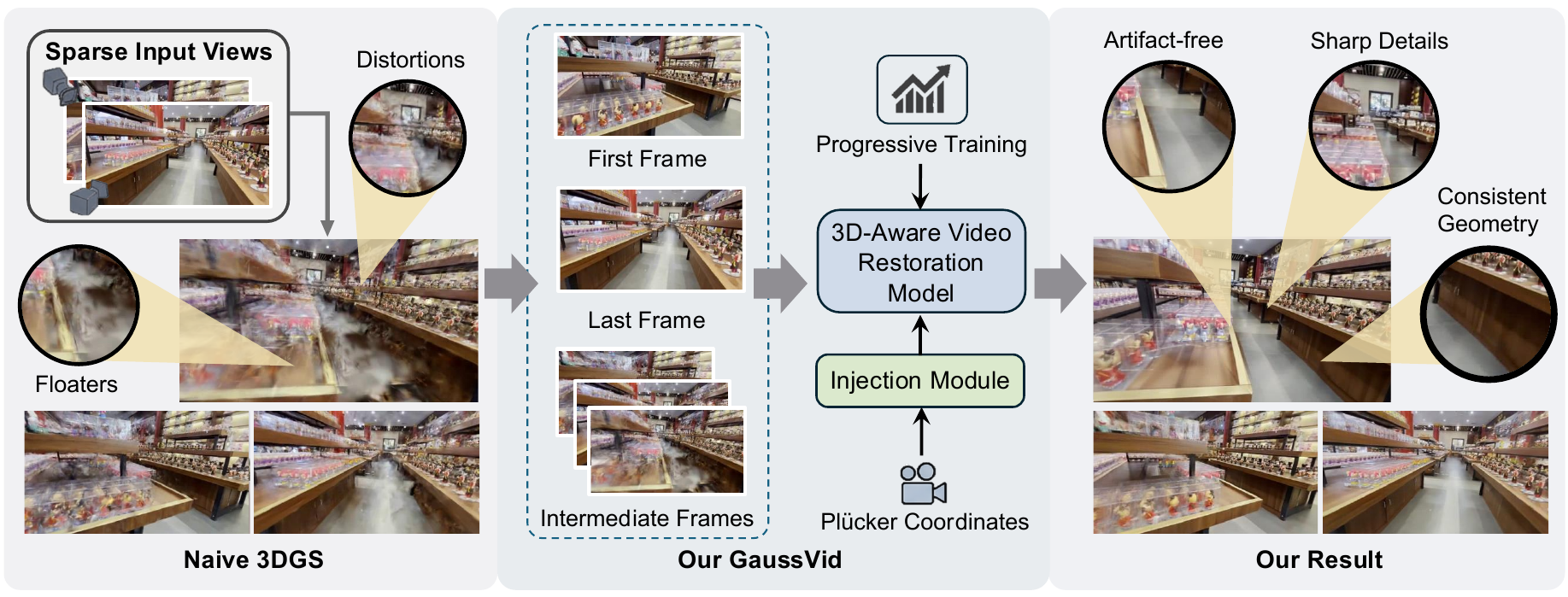}
  \caption{Comparison between naive 3DGS and our proposed GaussVid.
  Sparse 3DGS often produces floaters, distortions, and blurred structures.
  Our GaussVid restores intermediate frames using a first-to-last frame
  conditioned video model with a camera injection module and a progressive
  training, yielding artifact-free, sharp, and geometrically consistent results.}
  \label{fig:teaser}
\end{figure*}

\begin{abstract}
3D Gaussian Splatting (3DGS) has achieved remarkable success in novel view synthesis; however, reconstructions under sparse views often exhibit noticeable artifacts. While recent video diffusion models provide strong spatio-temporal priors for 3DGS restoration, directly fine-tuning them for restoration is suboptimal, as they lack awareness of the underlying multi-camera geometry, resulting in multi-view inconsistencies. In this work, we propose a novel 3D-aware video restoration framework designed to enhance the quality of sparse 3DGS reconstruction. Specifically, we construct a large-scale 3DGS video dataset to enable specialized fine-tuning. To bridge the gap between 2D video generation and 3D multi-view constraints, we introduce a camera-conditioned geometric prior. 
By using the first and last frames as boundary anchors and encoding the corresponding camera relationships, we explicitly inject spatial structure into the video generation pipeline. 
This boundary-anchored, camera-aware prior guides the network toward geometrically grounded restoration that remains coherent across viewpoints. Extensive experiments show that, among video-prior restoration methods, our approach attains the best pixel- and structure-level fidelity (PSNR/SSIM) and improves multi-view consistency, while remaining competitive in perceptual quality (LPIPS). 
\end{abstract}

\begin{IEEEkeywords}
Sparse-View Gaussian Splatting, Novel View Synthesis, Video Diffusion Models.
\end{IEEEkeywords}

\section{Introduction}
\IEEEPARstart{3}{D} Gaussian Splatting (3DGS)~\cite{kerbl20233d} has recently emerged as an explicit and efficient scene representation for novel view synthesis. 
While it demonstrates impressive performance under dense multi-view supervision, novel view synthesis from sparse observations remains a challenging problem.
The lack of sufficient multi-view constraints often leads to ambiguous geometry and overfitting, resulting in artifacts such as floaters, blurred structures, and poor generalization to unseen viewpoints~\cite{fei20243d,bao20253d,chen2024mvsplat,zhang2024cor}. 

Most existing approaches address this limitation by introducing various regularization strategies. Specifically, geometric priors, such as depth smoothness~\cite{zheng2025nexusgs,xiong2025sparsegs,paliwal2024coherentgs,patle2025ad,zhang2024cor,chung2024depth,wan2025s2gaussian,zhu2024fsgs} and normal consistency~\cite{xuenvc,li2025mpgs}, are imposed to encourage spatial coherence and suppress floaters. Meanwhile, sparsity and opacity regularization~\cite{park2025dropgaussian,li2024dngaussian,bao2025loopsparsegs,liu2024georgs,gu2024regsegfield} are employed to reduce redundancy and mitigate overfitting. In addition, appearance-level constraints, including perceptual and cross-view consistency losses~\cite{huang2025trigs,xiao2025mcgs,huang20253d}, are incorporated to improve generalization to unseen viewpoints. However, despite these efforts, such regularization-based methods remain sensitive to noise and typically provide limited improvements in novel view synthesis quality under sparse-view settings.

Recent works~\cite{wu2025difix3d,liu20243dgs,wang2024use,luo20253denhancer,liu2024deceptive,kong2025generative,zhu2026gaussfusionimproving3dreconstruction} attempt to address sparse-view ambiguity by incorporating 2D diffusion priors as additional image-level supervision. These approaches leverage pretrained diffusion models as powerful generative priors to guide rendered views toward more realistic and semantically consistent appearances. In practice, score distillation sampling (SDS)~\cite{pooledreamfusion,wang2023prolificdreamer} and progressive dataset updating techniques~\cite{haque2023instruct} are commonly adopted to align rendered images with the learned distribution of 2D diffusion models. Despite their effectiveness, such methods often suffer from optimization instability and cross-view inconsistency, as the underlying 2D priors lack explicit 3D awareness, which can result in geometry distortions and inconsistent details across viewpoints.

To overcome the limitations of 2D priors, more recent efforts explore the use of video diffusion models~\cite{yin2025gsfixer}, which inherently capture temporal consistency across frames. The key challenge lies in effectively aligning 2D temporal priors with the underlying 3D representation, enabling a 3D-aware video diffusion model that can provide temporally and geometrically consistent supervision.
An intuitive solution~\cite{yin2025gsfixer} is to leverage VGGT~\cite{wang2025vggt} to extract a coarse 3D prior from sparse input views, which can then serve as a geometric condition for training a video model.
However, the quality of this geometric condition is closely tied to the underlying estimation. While VGGT offers a useful coarse prior, the limited view overlap under sparse-view settings can leave the estimated geometry noisy or ambiguous. Consequently, the inaccurate geometry can provide somewhat weaker guidance to the diffusion process, leading to suboptimal supervision.

To address these challenges, we extend a first-to-last frame conditioned video model~\cite{wan2025wan} with 3D awareness, and exploit it as a prior to restore artifact-corrupted renderings from sparse 3DGS. 
We show that this first-to-last frame conditioned video model encodes visual and temporal consistency priors, which can be fine-tuned to correct artifacts in 3DGS renderings. 
To this end, we formulate sparse 3DGS restoration as a frame-conditioned reconstruction problem, where the first and last frames serve as anchors and the intermediate frames are treated as low-quality sources to be restored.
To enable supervised fine-tuning, we construct a large-scale paired dataset with diverse camera trajectories. 
To incorporate 3D awareness into the video prior, instead of conditioning on 3D structure that must be \emph{estimated} from the sparse views, we condition on the camera poses themselves---which are \emph{known} directly from the capture and rendering process---encoded as camera-ray (Pl\"{u}cker) features through a camera injection module. 
As a given rather than estimated quantity, these poses provide a reliable, noise-free geometric prior that does not degrade as the views become sparser. 
Furthermore, we propose a progressive training strategy to stabilize adaptation and enhance restoration performance.

Our main contributions are summarized as follows:
\begin{itemize}
    \item We propose a novel formulation of sparse 3DGS restoration as a frame-conditioned reconstruction problem, and construct a large-scale paired dataset with diverse camera trajectories to enable supervised fine-tuning.
    \item We extend a first-to-last frame conditioned video diffusion model with 3D awareness by introducing a camera injection module, and leverage it as a powerful prior to correct artifacts in sparse 3DGS, improving both visual fidelity and cross-view consistency. We further propose a progressive training strategy to stabilize fine-tuning and enhance restoration performance.
    \item Extensive experiments demonstrate that our method significantly outperforms existing approaches. Moreover, it serves as a general enhancement framework that can be seamlessly integrated to improve existing sparse 3DGS methods, and its restored frames can be fed back to refine the 3DGS itself (See the supplementary material).
\end{itemize}

\section{Related Work}
\subsection{Regularization for Sparse View 3DGS}
Most methods introduce various regularization strategies to impose geometric and appearance priors, such as depth smoothness~\cite{zheng2025nexusgs,xiong2025sparsegs,paliwal2024coherentgs,zhu2024fsgs}, normal consistency~\cite{xuenvc,li2025mpgs}, sparsity and opacity constraints~\cite{park2025dropgaussian,li2024dngaussian}, cross-view consistency and perceptual losses~\cite{huang2025trigs,xiao2025mcgs,huang2025structgs}. For example, FSGS~\cite{zhu2024fsgs} leverages a pretrained monocular depth estimator to provide depth-guided regularization for sparse-view 3DGS, encouraging accurate geometry growth from minimal inputs, while CoR‑GS~\cite{zhang2024cor} introduces a co‑regularization paradigm that jointly trains dual 3D Gaussian fields to detect and suppress inconsistent geometry, offering structural consistency through model-level mutual supervision.
Despite these efforts, regularization-based methods are often sensitive to noise and provide limited improvements under extreme sparse-view settings, leaving ambiguities in geometry and artifacts in rendered views.
In contrast, our method leverages a 3D-aware video diffusion prior that encodes temporal and spatial consistency, enabling robust restoration. This allows us to significantly reduce artifacts such as floaters and blurred structures without relying solely on handcrafted regularization terms.
\vspace{-3pt}

\subsection{2D Diffusion Priors for Sparse View 3DGS}
To overcome the limitations of handcrafted regularizations, recent works leverage pretrained 2D diffusion models as generative priors~\cite{wu2025difix3d,liu20243dgs,wang2024use,nguyen2026dediff,pan2026dreamjourney}. These approaches guide rendered views toward more realistic and semantically consistent appearances, typically using techniques such as SDS~\cite{pooledreamfusion,wang2023prolificdreamer} or progressive dataset updating~\cite{haque2023instruct}.
By aligning renderings with the learned distribution of large-scale 2D image models, diffusion priors can resolve ambiguities that regularization alone cannot. 
For example, Difix3D+~\cite{wu2025difix3d} introduces a single-step image diffusion model that enhances rendered novel views by removing artifacts and underconstrained regions, and then uses progressive dataset updating to distill these enhanced views back into the 3D representation, improving overall reconstruction quality.
However, the 2D nature of these priors limits their cross view consistency, often resulting in geometric distortions and inconsistent details across viewpoints.
By integrating a 3D-aware video diffusion prior, our approach overcomes the view-inconsistency issues of per-view diffusion enhancements. This yields more geometrically faithful reconstructions with improved cross-view consistency compared to single-step and iterative 2D diffusion methods.

\subsection{Video Priors for Sparse View 3DGS}
Extending beyond 2D priors, video diffusion models encode temporal consistency and spatial correlations across multiple frames, offering stronger supervision for sparse-view 3DGS. 
For instance, 4DSloMo~\cite{chen20254dslomo} adapts an image-to-video diffusion model via LoRA fine-tuning to generate intermediate views for sparse-view augmentation. While effective in improving visual continuity, this approach lacks explicit 3D awareness, making it difficult to align 2D temporal priors with the underlying 3D representation. As a result, the generated supervision may still suffer from geometric inconsistency across viewpoints.
To address this, GSFIXER~\cite{yin2025gsfixer} leverages VGGT~\cite{wang2025vggt} to extract a coarse 3D prior from sparse input views, which is then used as geometric conditioning for video diffusion. Although this strategy introduces a form of 3D guidance, its effectiveness is constrained by the quality of the estimated geometry. Under sparse-view settings, VGGT often produces noisy and ambiguous structures, which can mislead the diffusion process and lead to suboptimal supervision.
In contrast, we reformulate sparse 3DGS restoration as a frame-conditioned reconstruction problem, where the video model is guided by anchor frames to restore intermediate views. To explicitly incorporate 3D awareness, we propose a novel camera injection module, enabling geometry-consistent generation across viewpoints.
\vspace{-5pt}

\section{Method}
\label{sec:method}
\subsection{Overview}
Given a sparse set of input views 
$\mathcal{V} = \{V_1, \dots, V_N\}$ 
with corresponding camera poses 
$\mathcal{C} = \{c_1, \dots, c_N\}$, 
we first reconstruct a low-quality 3D scene using a 3D Gaussian Splatting model $G_\theta$, producing rendered frames 
$\hat{\mathcal{I}} = \{\hat{I}_1, \dots, \hat{I}_M\}$ 
with associated camera poses 
$\hat{\mathcal{C}} = \{\hat{c}_1, \dots, \hat{c}_M\}$, 
which often contain artifacts due to the limited multi-view observations. 
To restore the intermediate frames between two selected anchor views $(I_s, c_s)$ and $(I_e, c_e)$, we render $k$ frames from the sparse 3DGS, denoted as 
$(\hat{\mathcal{F}}, \hat{\mathcal{C}}_\mathcal{F}) = \{(\hat{F}_1, \hat{c}_1), \dots, (\hat{F}_k, \hat{c}_k)\}$. 
The restoration task is then formulated as recovering high-quality frames 
$\mathcal{F}^* = \{F_1^*, \dots, F_k^*\}$ 
from these low-quality sources via a 3D-aware video restoration model $F_\phi$ conditioned on the anchor frames, \emph{i.e.}, 
$$
\mathcal{F}^* = F_\phi((I_s, c_s), (I_e, c_e), (\hat{\mathcal{F}}, \hat{\mathcal{C}}_\mathcal{F})).
$$
The restored frames are expected to correct artifacts in the sparse 3DGS renderings while preserving temporal coherence, cross-view consistency, and geometric fidelity, thereby formalizing sparse 3DGS restoration as a frame-conditioned reconstruction problem with full camera awareness.

Our method consists of three key components.
\textbf{First, }since no existing dataset provides paired artifact-corrupted and clean video clips with associated camera poses, we first construct a large-scale training dataset for supervising the restoration model $F_\phi$ by simulating realistic sparse-view artifacts via controlled 3DGS optimization (Sec.~\ref{sec:data_curation}).
\textbf{Second}, to endow the video restoration model with 3D awareness, we introduce a camera-conditioning mechanism consisting of a Pl\"{u}cker Token Encoder, which encodes dense camera ray fields into geometric tokens, and a Dual-Stream Camera Injection scheme, which integrates these tokens into each DiT block via asymmetric attention with noise-invariant modulation (Sec.~\ref{sec:plucker_encoder}).
\textbf{Third}, we propose a progressive training strategy (Sec.~\ref{sec:training}) that combines a progressive latent reconstruction loss with a \emph{curriculum learning} schedule, gradually exposing the model to increasingly challenging samples to stabilize and enhance the fine-tuning process.
\vspace{-4pt}

\begin{figure*}[h]
\centering
\includegraphics[width=0.94\linewidth]{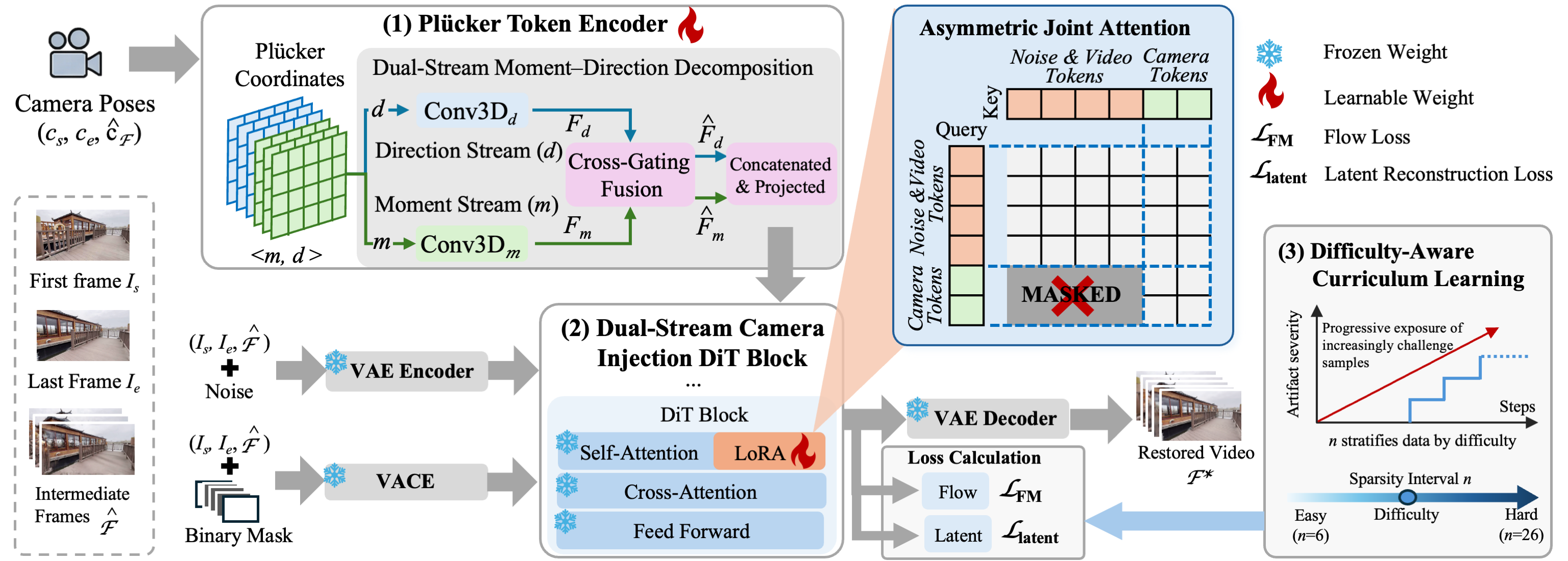}
\caption{Overview of the GaussVid framework. Our approach restores artifact-corrupted sparse 3DGS renderings into high-quality, 3D-consistent videos via three core designs: \textbf{(1) Pl\"{u}cker Token Encoder} extracts robust geometric priors by disentangling moment and direction streams via cross-gating fusion (Sec.~\ref{sec:Camera_Representation}). \textbf{(2) Dual-Stream Camera Injection} integrates these priors into DiT blocks using asymmetric joint attention, allowing video latents to query 3D context while shielding camera tokens from noise (Sec.~\ref{sec:dual_stream_dit}). \textbf{(3) Difficulty-Aware Curriculum Learning}: dynamically balances flow matching ($\mathcal{L}_{\text{FM}}$) and latent reconstruction ($\mathcal{L}_{\text{latent}}$) losses based on sample sparsity to stabilize optimization (Sec.~\ref{sec:curriculum}). \textit{Note: Snowflake icons denote frozen pre-trained base weights. Standard parameter-efficient fine-tuning details are provided in Sec.~\ref{sec:impl_details}.}}
\label{fig:pipeline}
\end{figure*}

\subsection{GaussVid Dataset Construction}
\label{sec:data_curation}
Training the restoration model $F_\phi$ requires paired artifact-corrupted and clean video sequences with associated camera poses. 
Since no existing dataset provides such data, we construct a large-scale training set from approximately 5{,}000 scenes in DL3DV~\cite{ling2024dl3dv}, sampling one 5-frame clip per scene.
We construct the dataset in two stages:

\paragraph{Artifact Simulation via Controlled Optimization.}
Formally, given a set of frames for a scene, we subsample sparse input views 
$\mathcal{V} = \{V_1, \dots, V_N\}$ 
at a random interval $n \in \{6, 8, 10, \dots, 26\}$, with corresponding camera poses 
$\mathcal{C} = \{c_1, \dots, c_N\}$. 
The remaining frames are reserved as held-out targets. 
We then reconstruct a low-quality 3D scene using a 3D Gaussian Splatting model $G_\theta$, intentionally terminating the optimization early at 8{,}000 iterations. In sparse-view settings, exhaustive optimization often leads to severe overfitting to the few available views, manifesting as extreme geometric distortions. By controlling the optimization steps, we capture the natural progression of these sparse-view artifacts ($e.g.$, floaters and blur) before the model degenerates entirely into view-overfitting.
Rendering the held-out viewpoints through the underfit model produces artifact-corrupted frames 
$\hat{\mathcal{I}} = \{\hat{I}_1, \dots, \hat{I}_M\}$ 
with corresponding camera poses 
$\hat{\mathcal{C}} = \{ \hat{c}_1, \dots, \hat{c}_M \}$, 
naturally paired with the original clean frames 
$\mathcal{I} = \{I_1, \dots, I_M\}$ 
and their camera poses 
$\mathcal{C}^* = \{c^*_1, \dots, c^*_M\}$. 
The sparsity interval $n$ directly controls the severity of artifacts, with larger $n$ yielding sparser views and more pronounced distortions, including floaters, blur, and structural inconsistencies.

\paragraph{Anchor-Based Clip Sampling.}
To support the frame-conditioned restoration, from each scene we sample a single 5-frame clip at $560 \times 1024$ resolution using a clean--noisy--clean strategy.
Let $(I_s, c_s)$ and $(I_e, c_e)$ denote the first and last anchor frames along with their camera poses, drawn from clean input views, and let 
$(\hat{\mathcal{F}}, \hat{\mathcal{C}}_\mathcal{F}) = \{(\hat{F}_1, \hat{c}_1), (\hat{F}_2, \hat{c}_2), (\hat{F}_3, \hat{c}_3)\}$ 
be the intermediate frames with their corresponding camera poses sampled from the artifact-corrupted renderings. Each training sample is thus formulated as
$$
(I_s, c_s, I_e, c_e, \hat{\mathcal{F}}, \hat{\mathcal{C}}_\mathcal{F}, \mathcal{F}^*),
$$
where 
$\mathcal{F}^* = \{F_1^*, F_2^*, F_3^*\}$  
are the high-quality targets for the intermediate frames. This setup provides temporal anchors and reliable references with full camera information for supervising the restoration model. 
In total, we obtain 5{,}000 training clips (one per scene) and 500 evaluation clips, each annotated with camera poses and the sparsity interval $n$, which serves as a controllable difficulty measure for curriculum learning (Sec.~\ref{sec:curriculum}).

\subsection{3D-Aware Video Restoration Model}
\label{sec:plucker_encoder}
Our method builds upon Wan2.1-VACE~\cite{wan2025wan}, a first-to-last frame conditioned video model. Given rendered video frames 
$(\hat{\mathcal{F}}, \hat{\mathcal{C}}_\mathcal{F}) \in \mathbb{R}^{C \times T \times H \times W}$ 
from the sparse 3DGS, Wan2.1-VACE decomposes the input into an \emph{inactive} component
$
\hat{\mathcal{F}}^{\text{inactive}} = \hat{\mathcal{F}} \odot (1 - Mask)
$
and a \emph{reactive} component
$
\hat{\mathcal{F}}^{\text{reactive}} = \hat{\mathcal{F}} \odot Mask,
$
where $Mask \in \{0,1\}^{1 \times T \times H \times W}$ is a binary mask. Both components are encoded by a frozen VAE and concatenated with a downsampled mask to form a context signal, which a lightweight VACE encoder converts into layer-wise hints additively injected into designated DiT blocks.
In our sparse 3DGS restoration setting, $Mask = 1$ for the intermediate frames $\hat{\mathcal{F}}$ and $Mask = 0$ for the anchor frames $(I_s, I_e)$.

However, such a model does not explicitly enforce 3D awareness, often leading to geometry inconsistencies and view-dependent artifacts across frames. To address this, we introduce explicit camera conditioning into the video model. Importantly, unlike prior work that conditions on 3D geometry \emph{estimated} from the sparse views ($e.g.$, via VGGT)---which is noisy and ambiguous under limited view overlap---we condition on the camera poses, which are \emph{given} directly by the rendering process. The poses thus serve as a reliable, noise-free geometric signal that does not rely on any error-prone estimation.
Existing camera-conditioned video generation methods~\cite{he2024cameractrl, zheng2024cami2v} typically inject camera geometry by directly adding Pl\"{u}cker ray maps to video latents, thereby entangling geometric and visual information from the outset. This early fusion makes it difficult for the network to disentangle camera priors from noisy, artifact-corrupted content. 
To address this, we introduce a novel camera representation based on Pl\"{u}cker coordinates to obtain geometry-aware features. We then design a dual-stream camera injection module that incorporates camera geometry as a \textit{separate token stream}, enabling it to interact with video tokens through attention. This design preserves the geometric identity of the camera throughout the network, facilitating more effective geometry-aware reasoning.

\subsubsection{Camera Representation}
\label{sec:Camera_Representation}
To incorporate explicit geometric cues, we associate each frame with its camera poses $(c_s, c_e, \hat{\mathcal{C}}_\mathcal{F})$
and map each pixel $(u_i, v_i)$ to a Pl\"{u}cker ray 
$r_i = \langle m_i, d_i \rangle \in \mathbb{R}^6$, 
where the direction is given by 
$d_i = R^{-1} K^{-1}(u_i, v_i, 1)^\top$,  
and the moment is defined as 
$m_i = C \times d_i$, 
with camera center $C = -R^{-1}T$. 
All camera poses are normalized to the first camera’s coordinate frame to ensure consistency across views.

\paragraph{Temporal Alignment.}
To enable effective interaction between camera and video representations, the camera features must be temporally aligned with the video latents. In Wan2.1-VACE, the input video is first temporally packed by repeating the first frame to form $T_{\text{packed}} = T + k$ frames, and then encoded by a 3D VAE to produce $F_{\text{latent}} = T_{\text{packed}} / r$ latent frames. We mirror this packing and encoding for the camera representation using the same temporal kernel $k$ and stride $r$, yielding camera tokens with exactly $F_{\text{latent}}$ temporal positions that correspond one-to-one with video latent tokens. This alignment ensures that camera-aware features can be injected into the video model in a temporally consistent manner.

\paragraph{Dual-Stream Moment–Direction Decomposition.} 
The Pl\"{u}cker ray encodes two fundamentally different geometric quantities: the moment $m_i$ captures the perpendicular distance from the origin to the ray, which is sensitive to translation, while the direction $d_i$ encodes the viewing angle, governed by rotation and camera intrinsics. Directly applying a convolution to the concatenated 6-channel field would entangle these distinct properties. To avoid this, we decompose the Pl\"{u}cker field into a moment stream $m \in \mathbb{R}^{3 \times T_{\text{packed}} \times H \times W}$ and a direction stream $d \in \mathbb{R}^{3 \times T_{\text{packed}} \times H \times W}$, each processed by a dedicated 3D convolutional patch embedding:
\begin{equation}
    F_m = \text{Conv3D}_m(m), \quad F_d = \text{Conv3D}_d(d).
\end{equation} 
\paragraph{Cross-Gating Fusion.}
We then fuse them via cross-gating: 
\begin{equation}
    \hat{F}_m = F_m \odot \sigma\!\big(g_{d \to m}(F_d)\big), \quad
    \hat{F}_d = F_d \odot \sigma\!\big(g_{m \to d}(F_m)\big),
\end{equation}
where $g_{d \to m}$ and $g_{m \to d}$ are lightweight MLPs, $\sigma$ is the sigmoid function, and $\odot$ denotes element-wise multiplication. The gated features are concatenated and projected to the DiT hidden dimension $D$:
\begin{equation}\label{eq:camera_proj}
    x_{\text{c}} = \gamma \cdot \text{LayerNorm}\!\big(\text{MLP}([\hat{F}_m;\, \hat{F}_d])\big) \in \mathbb{R}^{B \times N_{\text{c}} \times D},
\end{equation}
where $N_{\text{c}}$ is the number of camera tokens, matching the number of video tokens $N_v$. The final projection is zero-initialized and $\gamma$ is a learnable scalar initialized to $0.1$, ensuring that camera contributions are negligible at training start and the pretrained DiT behavior is preserved.

\subsubsection{Dual-Stream Camera Injection}
\label{sec:dual_stream_dit}
After obtaining the camera feature $x_c$, we now consider how to inject it into the video model. As previously mentioned, simply adding camera tokens to video latents merges geometric information with the noisy video representation, entangling it with artifact patterns. Na\"{i}ve concatenation further allows noisy video features to corrupt camera tokens through bidirectional attention. Drawing inspiration from the dual-stream mechanism in EasyControl~\cite{zhang2025easycontrol}, we design a camera injection scheme based on two guiding principles: (1) camera geometry serves as a noise-free prior and should be modulated independently of the diffusion timestep, and (2) the information flow must be asymmetric—video tokens absorb geometric cues, while camera tokens remain shielded from noisy video content.

\paragraph{Noise-Invariant Modulation.}  
Let  \( x_t \in \mathbb{R}^{B \times t \times N_v \times D} \) be the video token sequence at timestep $t$. In each DiT block, both video and camera tokens undergo LayerNorm and adaptive modulation.
For video tokens: $\hat{x}_t = (1 + s_v) \cdot \text{LN}(x_t) + b_v,$
while for camera tokens: $\hat{x}_c = (1 + s_c) \cdot \text{LN}(x_c) + b_c.$ Here, \( s_v \) and \( b_v \) are the scale and bias parameters for video tokens, and \( s_c \) and \( b_c \) are those for camera tokens. The timestep modulation embedding for video tokens is based on the current timestep \( t \), while for camera tokens, it is based on a fixed zero timestep.

\paragraph{Asymmetric Joint Attention.}  
The modulated tokens are then injected by a self-attention mechanism:
\begin{equation}
[\tilde{x}_t;\, \tilde{x}_c] = \text{SelfAttn} \!\big([\hat{x}_t;\, \hat{x}_c],\; M\big),
\end{equation}
where \( M \) is an asymmetric mask :
\begin{equation}
\resizebox{\columnwidth}{!}{$\displaystyle
M_{ij} =
\begin{cases}
0, & \text{if } i < N_v \quad \text{(video tokens attend to all tokens)}, \\
0, & \text{if } i \geq N_v \text{ and } j \geq N_v \quad \text{(camera$\to$camera)}, \\
-\infty, & \text{if } i \geq N_v \text{ and } j < N_v \quad \text{(camera$\not\to$video)}.
\end{cases}$}
\end{equation}
This allows video tokens to attend to all tokens, absorbing geometric context from camera tokens, while camera tokens only attend to other camera tokens, shielded from noisy video features.
\vspace{-3pt}

\subsection{Training Strategy}
\label{sec:training}
Our method is optimized within the flow matching framework~\cite{lipman2023flow}. Let $\mathbf{z}_0^*$ denote the latent representation of the high-quality target frames $\mathcal{F}^*$ encoded by the frozen VAE, and let $\hat{\mathbf{z}}$ denote the latent of the artifact-corrupted renderings $\hat{\mathcal{F}}$.
Following the observation that artifact-laden renderings approximate a partially noised version of the clean video~\cite{wu2025difix3d}, we treat the corrupted rendering as a mildly perturbed clean latent, $\hat{\mathbf{z}} \approx \mathbf{z}_0^* + \boldsymbol{\delta}$, where $\boldsymbol{\delta}$ denotes a low-magnitude, artifact-induced residual. Under this assumption, we construct the noisy latent $\mathbf{x}_t$ by interpolating between Gaussian noise $\boldsymbol{\epsilon} \sim \mathcal{N}(\mathbf{0}, \mathbf{I})$ and the corrupted rendering latent $\hat{\mathbf{z}}$, rather than the clean target:
\begin{equation}\label{noiseadd}
\mathbf{x}_t = (1 - t)\hat{\mathbf{z}} + t\boldsymbol{\epsilon}, 
\end{equation} 
so that no noise is injected at $t=0$ and the perturbation grows monotonically with $t$.
The restoration model $F_\phi$ is trained to predict the velocity field $\mathbf{v}_\phi$ that points toward the clean latent $\mathbf{z}_0^*$. The training objective is formulated as:
\begin{equation}\label{flowloss}
\mathcal{L}_{\text{FM}} = \mathbb{E}_{\mathbf{z}_0^*, t,\boldsymbol{\epsilon}} \left[ \left| (\boldsymbol{\epsilon} - \mathbf{z}_0^* ) - \mathbf{v}_\phi(\mathbf{x}_t; \text{c}, t) \right|_2^2 \right],
\end{equation}
where $\text{c}$ represents the joint conditioning signal comprising the VACE context (anchor frames $I_s, I_e$ and renderings $\hat{\mathcal{F}}$) and the geometric camera tokens $x_c$. We restrict training to the timestep interval $t \in [0.0, 0.48]$, ensuring the model operates in a noise regime consistent with the structural severity of 3DGS artifacts.
Moreover, we introduce two additional
components to stabilize and improve fine-tuning.

\paragraph{Progressive Latent Reconstruction Loss.}
While $\mathcal{L}_{\text{FM}}$ supervises the velocity field, it does not directly constrain the recovered clean output. We form an approximate clean-latent estimate $\hat{\mathbf{z}}_0 = \mathbf{x}_t - t \cdot \mathbf{v}_\phi$ and apply an auxiliary $L_1$ loss:
\begin{equation}\label{latentloss}
    \mathcal{L}_{\text{latent}} = \| \hat{\mathbf{z}}_0 - \mathbf{z}_0^* \|_1.
\end{equation}
Since the forward path is anchored at the corrupted latent $\hat{\mathbf{z}}$ rather than $\mathbf{z}_0^*$, $\hat{\mathbf{z}}_0$ is an approximate one-step reconstruction; the $L_1$ term drives it toward the clean target and is most reliable at low noise levels.
To stabilize training, we use a progressive weighting scheme:
\begin{equation}
    w_{\text{lat}} = \alpha \cdot \text{clamp}\!\left(\frac{s - S_w}{S_w}, 0, 1\right) \cdot \max\!\left(0,\, 1 - \frac{t}{\tau_p}\right),
\end{equation}
where $s$ is the current step, $S_w$ is the warm-up count, and $\tau_p$ is the timestep activation threshold. This ensures supervision is concentrated at lower noise levels, where the reconstruction is most accurate.

\paragraph{Difficulty-Aware Curriculum Learning.}
\label{sec:curriculum}
The subsampling interval $n \in \{6, \dots, 26\}$ naturally stratifies our training data by difficulty: a small interval ($n=6$) provides dense multi-view coverage with mild artifacts, whereas a large interval ($n=26$) introduces severe degradations and structural inconsistencies. Training on all difficulty levels simultaneously can lead to unstable optimization, as large, noisy gradients from high-difficulty samples may overwhelm the model before it learns basic reconstruction patterns.

To address this, we introduce a curriculum that progressively exposes the model to increasingly challenging samples. Let $s$ denote the current training step and $S$ the total number of steps. We define the normalized difficulty of a sample with interval $n$ as $\delta(n) = (n - n_{\min}) / (n_{\max} - n_{\min}) \in [0,1]$. A progress-dependent difficulty threshold $\delta_{\max}(s)$ is defined as:
\begin{equation}
    \delta_{\max}(s) = 
    \begin{cases} 
        0.2 \cdot \frac{s}{S \cdot p_w}, & s < S \cdot p_w \\
        0.2 + 0.8 \cdot \frac{s - S \cdot p_w}{S \cdot (1 - p_w)}, & s \geq S \cdot p_w
    \end{cases}
\end{equation}
where $p_w$ is the warm-up fraction. Samples exceeding this threshold ($\delta(n) > \delta_{\max}(s)$) are excluded from the current batch. For the remaining valid samples, we apply a soft weighting $w_{\mathrm{cur}}$ to prioritize easier cases while maintaining exposure to the full range of available data:
\begin{equation}
    w_{\mathrm{cur}}(n, s) = 
    \begin{cases} 
        1 - 0.4 \cdot \frac{\delta(n)}{\delta_{\max}(s)}, & \delta(n) \leq \delta_{\max}(s) \\
        0, & \delta(n) > \delta_{\max}(s)
    \end{cases}
\end{equation}
This schedule yields $w_{\mathrm{cur}} = 1$ for the simplest samples and decreases linearly to $w_{\mathrm{cur}} = 0.6$ at the difficulty boundary. This smooth transition, rather than a hard binary gate, stabilizes the fine-tuning of $F_\phi$ by ensuring the model masters fundamental geometric constraints before addressing sparse-view hallucination.

\paragraph{Total Objective.}
The final training objective for each sample is a weighted combination of the flow matching and latent reconstruction losses, modulated by the difficulty-aware curriculum:
\begin{equation}
    \mathcal{L} = w_{\mathrm{cur}}(n, s) \cdot \bigl(\mathcal{L}_{\text{FM}} + \lambda \cdot \mathcal{L}_{\text{latent}}\bigr),
\end{equation}
The curriculum weight $w_{\mathrm{cur}}$ modulates the entire objective, ensuring that gradients from high-difficulty samples only contribute to the optimization of $F_\phi$ after the model has developed sufficient reconstruction capacity from simpler instances. This staged supervision stabilizes the fine-tuning process, allowing the model to effectively bridge the gap between artifact-corrupted renderings and high-fidelity 3D-consistent videos.

\section{Experiments}
\subsection{Experimental Setup}
\label{sec:setup}
\textbf{Evaluation Benchmarks.}
We evaluate performance on two benchmarks to assess both in-distribution accuracy and cross-dataset generalization. (1) DL3DV~\cite{ling2024dl3dv}: We reserve 500 evaluation clips from scenes disjoint from the training set, rendered at intervals ($n \in {6, 8, 10,..., 24, 26}$) with a resolution of $560 \times 1024$, spanning various indoor and outdoor environments. (2) Zip-NeRF~\cite{barron2023zip}: To test generalization to unseen scenes and reconstruction methods, we evaluate four large-scale indoor scenes using Octree-GS~\cite{ren2025octree} as the base model. For each scene, we apply the same data construction pipeline at three sparsity levels ($n \in {6, 9, 12}$), producing 150 clips per sparsity level at a resolution of $576 \times 1024$, enabling evaluation across varying artifact severity.

\noindent\textbf{Baselines.}
We compare against three categories of methods:
(1) Optimization-based sparse 3DGS methods, including FSGS~\cite{zhu2024fsgs} and CoR-GS~\cite{zhang2024cor}.
(2) 2D diffusion prior-based methods, including DiFix3D~\cite{wu2025difix3d}.
(3) Video diffusion prior-based methods, including 4DSloMo~\cite{chen20254dslomo} and GSFIXER~\cite{yin2025gsfixer}.
For DiFix3D and 4DSloMo, we train each method on the same scenes as the vanilla 3DGS and extract the corresponding video clips for evaluation.
As GSFIXER releases only its inference code, we faithfully re-implement it following its original design---using VGGT~\cite{wang2025vggt} to extract 3D geometric features of the reference views and DINO-v2~\cite{oquab2023dinov2} to extract 2D semantic features---on the same Wan2.1-I2V-14B backbone as 4DSloMo, and train and evaluate it under the identical protocol for a fair comparison.

\noindent\textbf{Implementation Details.}
\label{sec:impl_details} 
We build upon the Wan2.1-VACE-14B model~\cite{wan2025wan} and inherit all default hyperparameters from its open-source release unless otherwise specified. 
Standard LoRA adapters (rank~32) are applied to the VACE module, while a tailored LoRA configuration is injected into the ${q, k, v, o, \text{ffn}}$ layers of the DiT module. The VAE encoder and decoder remain frozen. The Pl\"{u}cker Token Encoder is trained from scratch with full parameter updates. We use AdamW with a learning rate of $1 \times 10^{-4}$. Training clips consist of 5 frames at $560 \times 1024$ resolution. For the progressive latent reconstruction loss, we set $\lambda = 0.1$, $S_{w} = 1000$, and $\tau_{p} = 0.3$. For the difficulty-sensitive curriculum, the interval bounds are set at $n_{\min} = 6$ and $n_{\max} = 26$, with the total curriculum horizon at $S = 20,000$ steps and a warm-up fraction of $p_{w} = 0.05$. Samples whose difficulty exceeds the current threshold $\delta_{\max}(p)$ are skipped entirely ($w_{\min} = 0$). All experiments were conducted on 8 NVIDIA A800-SXM4-80GB GPUs. The project page is available at \url{https://github.com/Xinhui-99/GaussVid}.
\vspace{-6pt} 

\subsection{Comparison with Existing Methods}
\label{sec:comparison}
\begin{figure*}[htbp]
  \centering
  \includegraphics[width=0.96\linewidth]{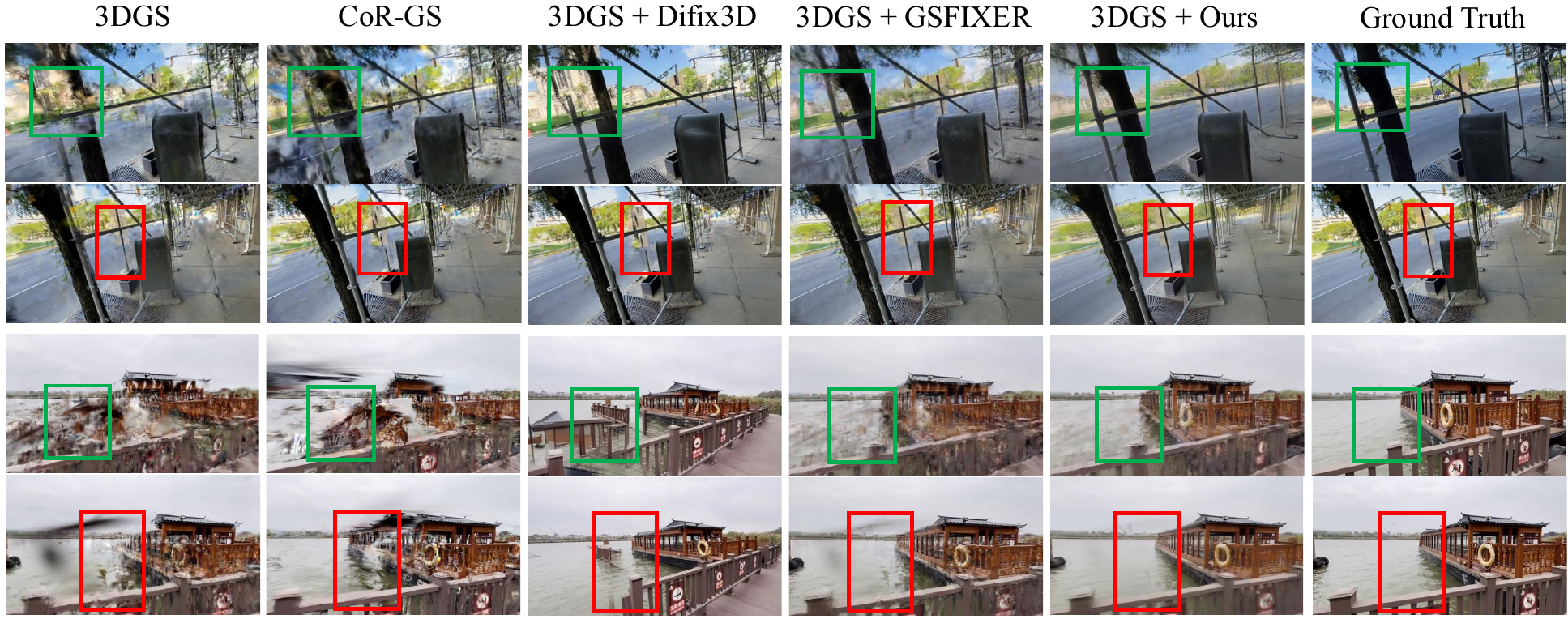}
  \caption{Qualitative comparison on DL3DV. From left to right: corrupted 3DGS~\cite{kerbl20233d}, CoR-GS~\cite{zhang2024cor}, 3DGS + DiFix3D~\cite{wu2025difix3d}, 3DGS + GSFIXER~\cite{yin2025gsfixer}, 3DGS + Ours, and Ground Truth. Colored insets highlight key differences: CoR-GS~\cite{zhang2024cor} still leave blurred boundaries in complex regions; DiFix3D shifts geometry and hallucinates content; although GSFIXER incorporates VGGT-derived 3D geometry, the noisy and ambiguous geometry under sparse views confines it to suboptimal results. Our method effectively removes floaters and structural distortions while preserving fine-grained details and multi-view consistency.}
  \label{fig:qual_dl3dv}
\end{figure*}

\noindent \textbf{Within-dataset evaluation.} Tab.~\ref{tab:generalization} presents a comprehensive comparison of the DL3DV benchmark. \textbf{First}, compared to the optimization-based sparse 3DGS methods such as FSGS and CoR-GS, our method consistently outperforms them, demonstrating its effectiveness without relying on manually designed, complex regularizations. \textbf{Second}, when considering 3DGS, FSGS, and CoR-GS as baselines and using DiFix3D, 4DSloMo, GSFIXER, and our method for restoration, our method achieves the best results in most cases. 
As a per-frame 2D generative model, DiFix3D attains the best LPIPS by synthesizing sharp, perceptually plausible textures that this metric rewards. However, such per-frame generation is not constrained across views: it shifts geometry and hallucinates content that deviates from the true scene (Fig.~\ref{fig:qual_dl3dv}), which lowers PSNR/SSIM and breaks multi-view consistency. Its LPIPS advantage therefore reflects single-frame perceptual sharpness rather than geometrically faithful reconstruction.
The video diffusion prior method 4DSloMo improves PSNR through temporal coherence but lacks explicit geometric guidance, limiting its gains in SSIM and LPIPS. 
GSFIXER, which incorporates VGGT-derived 3D geometry, achieves the second-best PSNR across all baselines; however, the geometry estimated under sparse views remains noisy and ambiguous, keeping it below our method, particularly in SSIM and LPIPS. 
In contrast, our method combines video-level consistency with camera-aware conditioning, achieving the best PSNR/SSIM and competitive LPIPS.
We further validate this consistency in the supplementary material, where the restored frames are fed back to re-optimize the sparse 3DGS, yielding a cleaner reconstruction and improved rendering on both DL3DV and MipNeRF-360.

\begin{table}[t] 
\centering
\caption{Comparison with existing restoration methods across different sparse-view 3DGS baselines on DL3DV. Each restoration method is applied as post-processing to the same set of baselines. Best results are in \textbf{bold}, second best are \underline{underlined}.}
\label{tab:generalization}
\begin{tabular}{l l c c c }
\hline
\textbf{Base } & \textbf{Restoration} & \textbf{PSNR$\uparrow$} &  \textbf{SSIM$\uparrow$} & \textbf{LPIPS$\downarrow$} \\
\hline
\multirow{4}{*}{3DGS~\cite{kerbl20233d}}
 & Baseline  & 20.67 & 0.7034 & 0.3815\\
 & + Difix3D~\cite{wu2025difix3d}  & 21.19 & 0.7221 & \textbf{0.2463}\\
 & + 4DSloMo~\cite{chen20254dslomo}  & 21.54 & 0.7052 & 0.3886  \\
 & + GSFIXER~\cite{yin2025gsfixer}  & \underline{22.35} & \underline{0.7239} & 0.3685  \\
 & + \textbf{Ours}  & \textbf{22.74} & \textbf{0.7291} & \underline{0.3599} \\
\hline
\multirow{4}{*}{FSGS~\cite{zhu2024fsgs}}
 & Baseline  & 21.13 & 0.7289 &  0.3620 \\
 & + Difix3D~\cite{wu2025difix3d}  & 20.93 & \underline{0.7357} & \textbf{0.3096} \\
 & + 4DSloMo~\cite{chen20254dslomo} & 21.00 & 0.7346 & 0.3748 \\
 & + GSFIXER~\cite{yin2025gsfixer}  & \underline{21.19} & 0.7306 &  0.3594  \\
 & + \textbf{Ours}  & \textbf{21.36} & \textbf{0.7377} & \underline{0.3563}  \\ 
\hline
\multirow{4}{*}{CoR-GS~\cite{zhang2024cor}}
 & Baseline   & 21.26 & 0.7306 & 0.3295\\
 & + Difix3D~\cite{wu2025difix3d}  & 21.51 & \underline{0.7382} & \textbf{0.2266} \\
 & + 4DSloMo~\cite{chen20254dslomo}  & 21.69 & 0.7273 & 0.3528  \\
 & + GSFIXER~\cite{yin2025gsfixer}   & \underline{22.17} & 0.7261 &  0.3273 \\
 & + \textbf{Ours} & \textbf{22.80} & \textbf{0.7474} & \underline{0.3244}  \\
\hline
\end{tabular}%
\end{table}

\vspace{-3pt}

Figure~\ref{fig:qual_dl3dv} presents qualitative comparisons on the DL3DV benchmark. The corrupted 3DGS renderings exhibit semi-transparent floaters and structural distortions (highlighted in green and red insets). CoR-GS~\cite{zhang2024cor} reduces some floaters but leaves blurred boundaries in complex regions. DiFix3D~\cite{wu2025difix3d} introduces geometric misalignment and hallucinated textures not found in the ground truth (green and red insets). 
GSFIXER~\cite{yin2025gsfixer} leverages VGGT-derived 3D geometry and removes most floaters, but the noisy and ambiguous geometry under sparse views leaves less faithful fine structures.
In contrast, our method effectively removes floating rocks, recovers sharp boundaries, and preserves fine details such as railing edges and wooden textures, yielding results closest to the ground truth.

\begin{figure*}[h]
  \centering
  \includegraphics[width=0.94\linewidth]{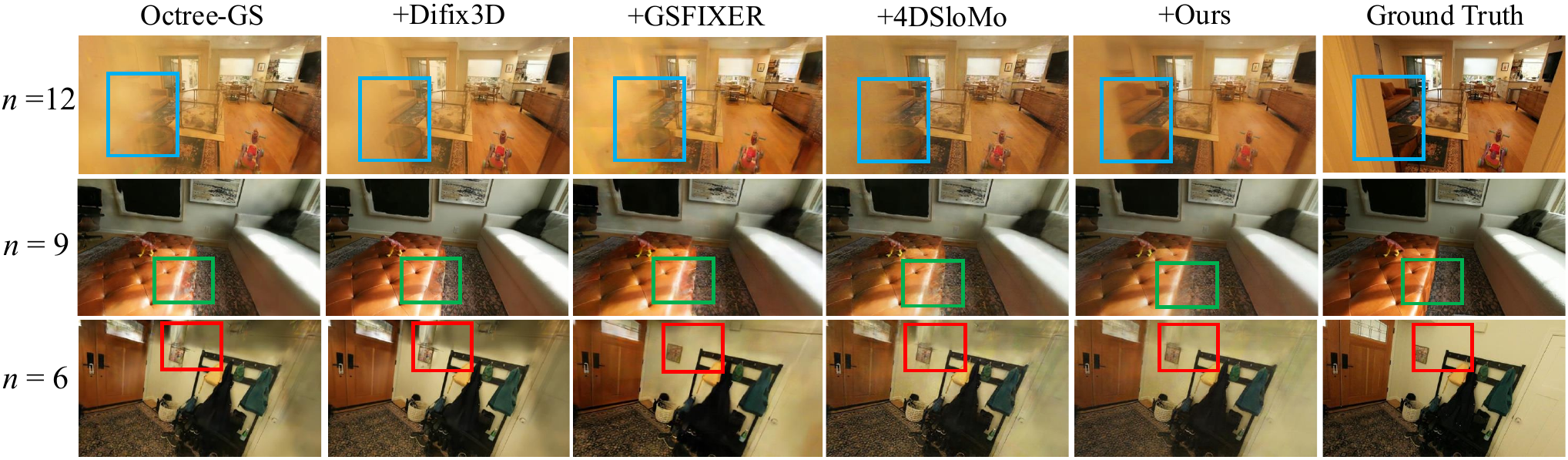}
  \caption{Qualitative comparison on Zip-NeRF~\cite{barron2023zip} indoor scenes at three sparsity levels ($n{=}6, 9, 12$). DiFix3D over-generates content misaligned with the ground truth (green and red insets); although GSFIXER leverages VGGT-derived 3D geometry, the noisy geometry under sparse views leaves less faithful fine structures; 4DSloMo retains residual local artifacts (blue and green insets). Our method yields the cleanest results across all settings.}
  \label{fig:cmp_zipnerf}
\end{figure*}

\noindent \textbf{Cross-dataset evaluation.} 
To evaluate our method across different capture setups and sparsity levels, we tested indoor scenes from Zip-NeRF~\cite{barron2023zip}, which differ significantly from DL3DV in terms of capture trajectory and scene distribution. We used Octree-GS~\cite{ren2025octree} as the base reconstruction method and evaluated on videos rendered with three training configurations ($n \in {6, 9, 12}$).
As shown in Tab.~\ref{tab:zipnerf}, our method outperforms others in PSNR across all sparsity levels. These results confirm the effectiveness of our method under varying capture conditions and sparsity levels.
Fig.~\ref{fig:cmp_zipnerf} presents qualitative comparisons. We observe similar results to those seen on the DL3DV benchmark. Our method consistently yields the best results, effectively removing artifacts and preserving fine details.

\begin{table}[t] 
\centering
\caption{Generalization to Zip-NeRF~\cite{barron2023zip} indoor scenes with Octree-GS~\cite{ren2025octree} as the base model. PSNR is reported at three sparsity levels ($n{=}6, 9, 12$). Best results in \textbf{bold}.}
\label{tab:zipnerf}
\small
\setlength{\tabcolsep}{12pt}  
\resizebox{\columnwidth}{!}{
\begin{tabular}{l c c c}
\hline
\textbf{Methods}  & $n$=6 & $n$=9  & $n$=12 \\
\hline
Baseline (Octree-GS~\cite{ren2025octree})  &21.02 & 20.62& 20.13\\
 + Difix3D~\cite{wu2025difix3d}& 21.20& 21.04 & 20.34 \\
 + 4DSloMo~\cite{chen20254dslomo}& 21.79& 21.43 & 20.98  \\
 + GSFIXER~\cite{yin2025gsfixer}&21.83 & 21.12 &  20.76 \\
 + Ours& \textbf{21.99} & \textbf{21.84} & \textbf{21.16} \\
\hline
\end{tabular}}
\end{table}
\vspace{-3pt} 

\vspace{-3pt}

\subsection{Ablation Studies}
\label{sec:ablation}
We perform ablation studies on the DL3DV benchmark using vanilla 3DGS as the base reconstruction method to validate each design choice. All ablation variants are trained under the same computational budget and evaluation protocol.

\subsubsection{Effect of Data Construction Pipeline}
\label{sec:Construction_Pipeline}
Our data construction pipeline involves three design choices: (a)~how many clean anchor frames to include, (b)~how to sample the intermediate corrupted frames, and (c)~what range of artifact difficulty to cover during training. We ablate each dimension independently in Tab.~\ref{tab:data_construction}, where every variant modifies exactly one setting from our full configuration while keeping the others fixed. 
\noindent\textbf{(a)} Clean anchor frames.
We vary the number of clean anchor frames per clip. Removing all anchors (21.90~dB) forces the model to restore without any reliable reference, while using only the first frame as anchor (22.11~dB) provides a one-sided constraint. Our dual-anchor design (22.74~dB) yields a +0.84~dB gain over the non-anchor variant, confirming that clean boundary frames at both ends effectively guide the restoration of intermediate corrupted frames.
\noindent\textbf{(b)} Intermediate frame sampling.
We replace our uniform sampling of the three intermediate corrupted frames with random selection. Random sampling (22.35~dB) underperforms uniform sampling (22.74~dB) by 0.39~dB, as evenly spaced frames provide more consistent temporal coverage along the camera trajectory.
\noindent\textbf{(c)} Artifact difficulty diversity.
We compare training on narrow-interval data only ($n{=}6$, mild artifacts), wide-interval data only ($n{=}26$, severe artifacts), and our full mixed range ($n \in \{6, 8, \dots, 26\}$). Narrow-only (22.19~dB) is less exposed to severe degradations; wide-only (22.47~dB) misses easy samples needed to establish stable restoration patterns. Our mixed strategy (22.74~dB) outperforms both, showing that a wide coverage of difficulties is essential for generalization.

\begin{table}[t]
\centering
\caption{\textbf{Ablation on data construction pipeline.} Each variant modifies exactly one design choice from our full configuration. (a)~Number of clean anchor frames. (b)~Intermediate frame sampling strategy. (c)~Artifact difficulty range during training. Best results in \textbf{bold}.}
\label{tab:data_construction}
\small
\setlength{\tabcolsep}{4.5pt}
\resizebox{\columnwidth}{!}{
\begin{tabular}{l l c c c}
\hline
 & \textbf{Variant} & \textbf{PSNR$\uparrow$} & \textbf{SSIM$\uparrow$} & \textbf{LPIPS$\downarrow$} \\
\hline
\multirow{2}{*}{\scriptsize(a)}
 & No anchor (all corrupted)         & 21.90 & 0.7160 & 0.3692 \\
 & First-only anchor                 & 22.11 & 0.7192 & 0.3679 \\ 
\hline
\multirow{1}{*}{\scriptsize(b)}
 & Random intermediate sampling      & 22.35 & 0.7240 & 0.3685 \\
\hline
\multirow{2}{*}{\scriptsize(c)}
 & Narrow-only ($n{=}6$)             & 22.19 & 0.7134 & 0.3767 \\
 & Wide-only ($n{=}26$)              & 22.47 & 0.7255 & 0.3642 \\
\hline
 & \textbf{Ours}                     & \textbf{22.74} & \textbf{0.7291} & \textbf{0.3599} \\
\hline
\end{tabular}}
\end{table}

\subsubsection{Effect of Camera Conditioning Injection}
\label{sec:abl_camera}
\begin{figure}[t]
  \centering
  \includegraphics[width=0.99\linewidth]{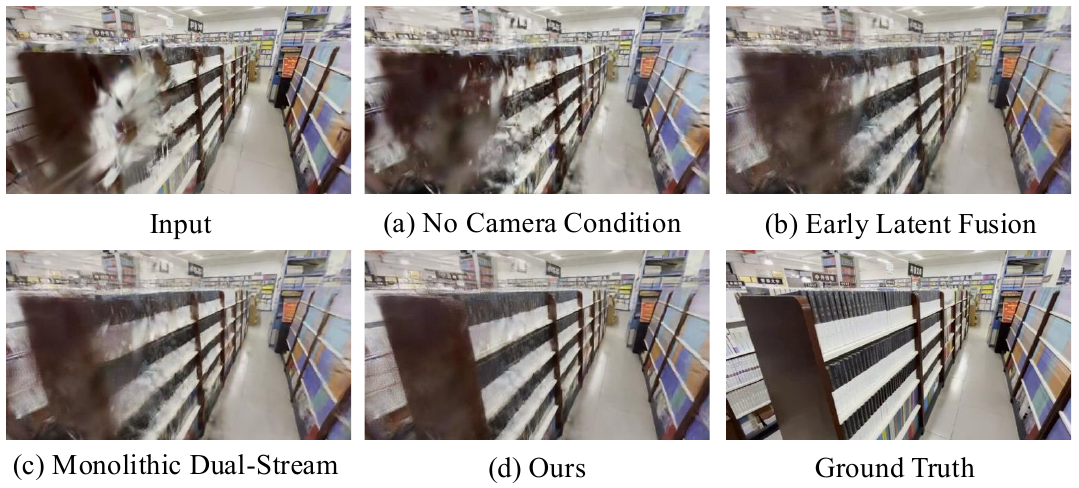}
\caption{Qualitative ablation on camera conditioning. (a)~No Camera Condition; (b)~Early Latent Fusion; (c)~monolithic dual-stream; (d)~our decomposed cross-gate injection. Our design recovers the sharpest geometry, especially near depth discontinuities (see shelf edges).}
  \label{fig:abl_camera}
\end{figure}

We ablate four camera conditioning strategies with increasing geometric expressiveness (Tab.~\ref{tab:camera_injection}, Fig.~\ref{fig:abl_camera}).
\noindent\textbf{(a) No Camera Condition.}
The model receives only the VACE context without any camera geometry, achieving only 21.27 \,dB, which confirms that explicit geometric awareness is indispensable.
\noindent\textbf{(b) Early Latent Fusion.}
Following~\cite{he2024cameractrl, zheng2024cami2v}, we add 6-channel Pl\"{u}cker ray maps directly to the video latents before the DiT blocks. This early entanglement of geometric and visual signals yields only a marginal gain (+0.44 \, dB), as the network struggles to disentangle camera priors from artifact-corrupted content.
\noindent\textbf{(c) Monolithic Dual-Stream.}
Inspired by EasyControl~\cite{zhang2025easycontrol}, camera geometry is maintained as a separate token stream with asymmetric attention, but the full 6-channel Pl\"{u}cker coordinates are encoded jointly by a single 3D convolution. This improves over~(a) by +0.82\,dB, validating the separate-stream design, yet the monolithic encoding conflates the physically distinct moment (translation-sensitive) and direction (rotation-sensitive) components, limiting geometric expressiveness.
\noindent\textbf{(d) Decomposed Cross-Gate Injection (Ours).}
Our design further decomposes the Pl\"{u}cker field into separate moment and direction streams with dedicated embeddings, fused via cross-gating before injection through noise-invariant asymmetric attention (Sec.~\ref{sec:plucker_encoder}). This achieves the best results for all metrics (+1.47 \, dB over~(a)), confirming that disentangling geometrically meaningful components enables more effective camera-aware reasoning. As shown in Fig.~\ref{fig:abl_camera}, our method recovers the sharpest geometry, especially near depth discontinuities such as shelf edges.

\begin{table}[t]  
\centering
\caption{\textbf{Ablation on camera conditioning injection.} We compare strategies of increasing sophistication for injecting camera geometry into the diffusion model. Best results in \textbf{bold}.}
\label{tab:camera_injection}
\small
\setlength{\tabcolsep}{6pt}
\resizebox{\columnwidth}{!}{
\begin{tabular}{l ccc}
\hline
\textbf{Injection Strategy} & \textbf{PSNR}$\uparrow$ & \textbf{SSIM}$\uparrow$ & \textbf{LPIPS}$\downarrow$ \\
\hline
(a) No Camera Condition                    & 21.27 & 0.7104 & 0.3805 \\
(b) Early Latent Fusion                & 21.71 & 0.7155 & 0.3800 \\
(c) Monolithic Dual-Stream             & 22.09 & 0.7179 & 0.3700 \\
(d) Ours      & \textbf{22.74} & \textbf{0.7291} & \textbf{0.3599} \\
\hline
\end{tabular}}
\end{table}

\subsubsection{Effect of Training Strategy}
\label{sec:abl_training}
\begin{figure}[t]
  \centering
  \includegraphics[width=0.99\linewidth]{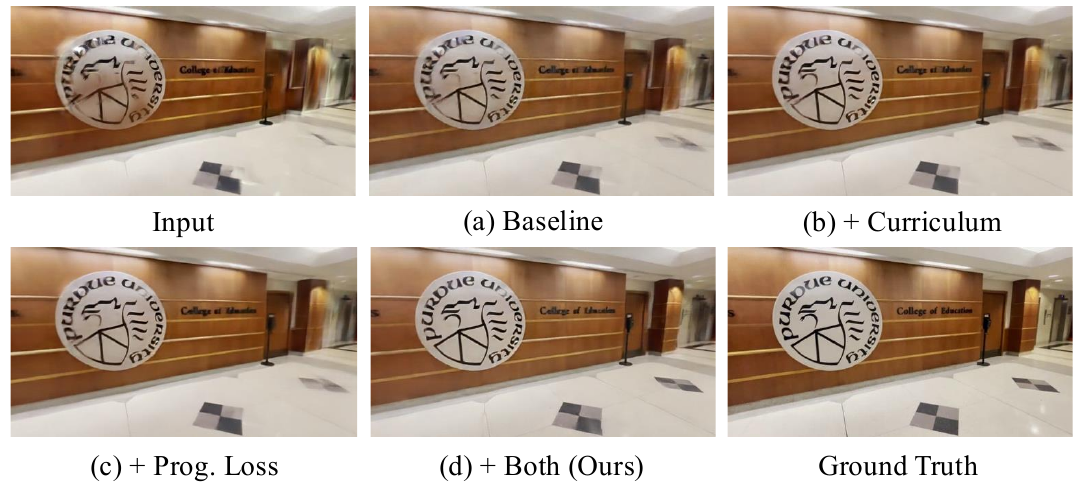}
  \caption{\textbf{Qualitative effect of each training component.} Between the corrupted input and the ground truth, we compare a shuffle-trained baseline~(a) with each component added incrementally: (b)~a difficulty-aware curriculum that gradually raises clip difficulty from easy to hard (small to large sampling interval $n$); (c)~a progressive latent-reconstruction loss that shifts supervision from latent space toward perceptual quality; and (d)~both combined (Ours). The curriculum mainly removes large-scale structural artifacts, the progressive loss restores fine-grained texture, and combining both yields the result closest to the ground truth.}
  \label{fig:training_strategy}
\end{figure}

We ablate two key components of our training strategy (Tab.~\ref{tab:training_strategy}, Fig.~\ref{fig:training_strategy}:) (1)~\textbf{Difficulty-Aware Curriculum Learning (Curriculum)}, which progressively increases clip difficulty from easy (small $n$, narrow camera intervals) to hard (large $n$, wide intervals); and (2)~\textbf{Progressive Latent Reconstruction Loss (Prog. Loss)}, which gradually transitions supervision from latent-space reconstruction to perceptual quality.
As shown in Tab.~\ref{tab:training_strategy}, each component yields consistent gains over the baseline trained with shuffle. Curriculum scheduling primarily suppresses large-scale structural artifacts by exposing the model to increasingly challenging viewpoint transitions, while progressive loss recovers fine-grained texture by steering optimization toward perceptual fidelity in later training stages (Fig.~\ref{fig:training_strategy}). In particular, the two designs are complementary -- one structures the difficulty of the data, the other structures the learning objective---and their combination achieves the best overall performance.  

\begin{table}[t] 
\centering
\caption{\textbf{Ablation on training strategy.} Each proposed component is incrementally added to the shuffle-trained baseline. All variants use identical data and total training steps. Best results in \textbf{bold}.}
\label{tab:training_strategy}
\small
\setlength{\tabcolsep}{9pt}
\vspace{-3pt}
\resizebox{\columnwidth}{!}{
\begin{tabular}{l ccc}
\hline
\textbf{Setting} & \textbf{PSNR$\uparrow$} & \textbf{SSIM$\uparrow$} & \textbf{LPIPS$\downarrow$} \\
\hline
(a) Baseline (Shuffle + $\mathcal{L}_\text{FM}$)
    & 21.50  & 0.7135  & 0.3828  \\
(b) +\,Curriculum 
    & 22.06 & 0.7134 & 0.3767 \\
(c) +\,Prog. Loss
    & 22.54 & 0.7261 & 0.3658 \\
(d) +\,Both (Ours) 
    & \textbf{22.74} & \textbf{0.7291} & \textbf{0.3599}  \\
\hline
\end{tabular}}
\end{table}

\begin{table}[t]
\centering
\caption{\textbf{Ablation on the adaptation strategy.} The backbone is fixed to Wan2.1-VACE-14B and only the adaptation strategy is varied. $^{*}$~applies the backbone zero-shot; $^{**}$~fine-tunes it on our data with vanilla LoRA, without our camera injection or progressive training. All variants share the same 3DGS baseline. Best results in \textbf{bold}.}
\label{tab:comparison}
\small
\setlength{\tabcolsep}{9pt}
\begin{tabular}{l ccc}
\hline
\textbf{Setting} & \textbf{PSNR$\uparrow$} & \textbf{SSIM$\uparrow$} & \textbf{LPIPS$\downarrow$} \\
\hline
 Baseline (3DGS)
    & 20.67 & 0.7034 & 0.3815 \\
(a) +\,Wan2.1$^{*}$
    & 20.56 & 0.6929 & 0.4050 \\
(b) +\,Wan2.1$^{**}$
    & 21.11 & 0.7071 & 0.3915 \\
(c) +\,Ours
    & \textbf{22.74} & \textbf{0.7291} & \textbf{0.3599} \\
\hline
\end{tabular}
\end{table}

\subsubsection{Effect of the Proposed Method over Naive Adaptation}
\label{sec:basemodel} 
We ablate three adaptation strategies of increasing capability under a fixed backbone\footnote{We use Wan2.1-VACE-14B; the recent video-prior baselines 4DSloMo and GSFIXER adopt the same-family Wan2.1-I2V-14B variant of comparable scale, and we choose the VACE variant because its first-to-last-frame conditioning is required by our frame-conditioned formulation (Sec.~\ref{sec:plucker_encoder}).} (Tab.~\ref{tab:comparison}), isolating our method's contribution from the backbone itself.
\noindent\textbf{(a) Zero-shot Backbone.}
The pre-trained backbone is applied directly without any fine-tuning, achieving only $20.56$\,dB---even below the 3DGS baseline ($20.67$\,dB)---as it lacks task-specific knowledge and 3D awareness.
\noindent\textbf{(b) Vanilla LoRA Fine-tuning.}
Fine-tuning the same backbone on our paired data with vanilla LoRA, but without our camera injection or progressive training, brings only a marginal $+0.44$\,dB gain over the baseline, indicating that naive adaptation is insufficient.
\noindent\textbf{(c) Full GaussVid (Ours).}
Equipping the identical backbone and data with our camera-aware conditioning and progressive training strategy delivers a substantial $+2.07$\,dB over the baseline ($+1.63$\,dB over~(b)), with consistent SSIM and LPIPS gains. Since all variants share the same backbone, these gains stem from our proposed method rather than the backbone itself.
\vspace{-3pt}  

\section{Conclusion}
We introduced \textbf{GaussVid}, a new framework that leverages 3D-aware video diffusion priors to restore artifact-corrupted renderings from sparse-view 3DGS. To tackle the challenge of multi-view consistent restoration, our approach makes three key contributions. First, we proposed a scalable paired-data construction pipeline via controlled 3DGS optimization, enabling the generation of realistic, varying-difficulty artifact data without manual annotation. Second, we designed a robust camera-aware conditioning mechanism that decomposes Pl\"{u}cker coordinates into distinct moment and direction streams, injecting them into the diffusion process through noise-invariant asymmetric attention to preserve geometric integrity. Finally, we introduced a difficulty-aware curriculum learning strategy coupled with a progressive latent reconstruction loss to stabilize the fine-tuning process. Extensive experiments on the DL3DV and Zip-NeRF benchmarks demonstrate that GaussVid effectively removes floating and structural distortions. It achieves state-of-the-art quantitative and qualitative performance, exhibiting strong generalization capabilities across diverse 3DGS baselines and unseen scenes.
 
\bibliographystyle{IEEEtran}
\bibliography{references}

\end{document}